\documentclass{article}

\usepackage[english]{babel}

\usepackage[a4paper, left=1.91cm, right=1.91cm, top=2.5cm, bottom=2.5cm]{geometry}
\usepackage{amsfonts}
\usepackage{tabularx}
\usepackage{multicol}
\usepackage{amsmath}
\usepackage{newtxtext}  
\usepackage{enumitem}
\usepackage{titlesec}
\usepackage{newtxmath}
\usepackage{booktabs}
\usepackage{graphicx}
\usepackage{caption}
\usepackage{float}
\usepackage[colorlinks=true, allcolors=blue]{hyperref}
\usepackage{natbib}
\usepackage{hyperref}
\usepackage[capitalise]{cleveref}
\title{Element-wise Modulation of Random Matrices for Efficient Neural Layers}
\author{Maksymilian Szorc}

\begin{document}

\begin{center}
\rule{\textwidth}{1.5pt}

\vspace{0.5cm}

\begin{minipage}{0.85\linewidth}
    \centering
    {\large \bfseries Element-wise Modulation of Random Matrices for Efficient Neural Layers}
\end{minipage}
\vspace{0.25cm}

\rule{\textwidth}{1pt}
\vspace{1.5cm}
\renewcommand{\thefootnote}{\fnsymbol{footnote}}
\setcounter{footnote}{0}

\begin{minipage}{0.9\textwidth}
\centering
\begin{tabular}[t]{c}
    \textbf{Maksymilian Szorc} \\
    Polish-Japanese Academy of Information Technology, Warsaw \\
    Universal Computing Research \\
    \texttt{\{s27456@pjwstk.edu.pl, mszorc@universalcomputingresearch.eu\}}
\end{tabular}
\end{minipage}
\vspace{2cm}
\end{center}

\begin{multicols}{2}

\begin{abstract}
Fully connected layers are a primary source of memory and computational overhead in deep neural networks due to their dense, often redundant parameterization. While various compression techniques exist, they frequently introduce complex engineering trade-offs or degrade model performance. We propose the Parametrized Random Projection (PRP) layer, a novel approach that decouples feature mixing from adaptation by utilizing a fixed random matrix modulated by lightweight, learnable element-wise parameters. This architecture drastically reduces the trainable parameter count to a linear scale while retaining reliable accuracy across various benchmarks. The design serves as a stable, computationally efficient solution for architectural scaling and deployment in resource-limited settings.
\end{abstract}

\section{Introduction}
Fully connected (FC) or projection layers are a core component of modern neural networks. Despite their generality and effectiveness, their dense parametrization incurs a prohibitive parameter cost, making them one of the main contributors to the computational and memory overhead of deep models. A layer mapping a 2048-dimensional input to a 1024-dimensional feature space, a common component in many models, comprises over 2 million trainable parameters.\medskip

Recent studies have increasingly questioned the necessity of dense parameterization, as many parameters appear redundant \citep{denil2014predictingparametersdeeplearning}. This has spurred exploration into alternatives, but they often involve trade-offs: rank-constrained decompositions \citep{Sainath2013LowrankMF} can degrade performance, post-training compression (like quantization \citep{gupta2015deeplearninglimitednumerical}) does not fundamentally alter the projection mechanism, and selective computation methods (like MoE \citep{shazeer2017outrageouslylargeneuralnetworks}) introduce complex mechanisms which can improve computational efficiency at the cost of added architectural and engineering complexity. A simple, general, and efficient alternative to dense projections remains an open problem.\medskip

Decades of research in high-dimensional statistics, physics, and signal processing shows that many complex mappings can be effectively captured using structured random transformations. Random projections provide a clear example, preserving essential geometric properties, like distances and angles, even when dimensionality is dramatically reduced.\medskip

In this work, we revisit classical ideas of random and structured projections through a modern deep learning lens. Traditional random projections are fixed, non-adaptive, and mainly serve to reduce the dimensionality of complex, high-dimensional data \citep{Bingham2001RandomPI}. This motivates us to propose the Parametrized Random Projection layers. The proposed layer performs a fixed random (or structured) projection from the input to the output space, while introducing learnable element-wise modulation parameters that adjust the scaling and bias of the features without altering the underlying projection structure. This design effectively decouples feature mixing and modulation. This results in a lightweight, stable, and expressive parameterization capable of learning effective linear mappings within the fixed random subspace with significantly fewer trainable parameters while being compatible with existing training strategies and models.

This approach leads to our main contributions:
\renewcommand{\labelitemi}{-}
\begin{itemize}[itemsep=0.5ex, leftmargin=*]
\item We propose and formalize the Parametrized Random Projection (PRP) layer, a drop-in replacement for standard fully connected layers that can project between arbitrary input and output dimensions. (\cref{sec:prp_layer})

\item Our results demonstrate that the PRP layer reduces trainable parameters from $d_{\text{in}} \cdot d_{\text{out}} + d_{\text{out}}$ in a standard layer to a more efficient $d_{\text{in}} + 2d_{\text{out}}$, while also reducing training overhead, specifically the costs associated with gradient computation and optimizer state. (\cref{sec:compute})

\item We empirically demonstrate that Multi-Layer Perceptrons (MLPs) and classifier replacements across popular models built with PRP layers achieve competitive accuracy across several benchmarks with a fraction of the trainable parameters of traditional dense layers. (\cref{sec:results})
\end{itemize}

\section{Parametrized Random Projection (PRP) Layer}

The Parametrized Random Projection (PRP) layer is a highly efficient alternative to the standard fully connected (FC) layer. The core intuition is to replace the single, large, trainable weight matrix $W$ with a decomposed structure. This new structure uses a fixed, non-trainable random matrix to perform the mixing and projection of features across dimensions. The layer's adaptability and expressive power then come from lightweight, learnable parameters that modulate the inputs and outputs of this fixed projection. 
\subsection{Proposed PRP Layer}
\label{sec:prp_layer}

The transformation is defined by the equation:
\begin{equation}
\mathbf{y}
=
\big( P^\top (\mathbf{x} \odot \boldsymbol{\alpha}) \big)
\odot \mathbf{w}
+ \mathbf{b},
\end{equation}
where $\odot$ denotes elementwise multiplication.\medskip

Learnable parameters: 
$\boldsymbol{\alpha}\in\mathbb{R}^{d_{\mathrm{in}}}$ (input scaling), 
$\mathbf{w}\in\mathbb{R}^{d_{\mathrm{out}}}$ (output scaling), and 
$\mathbf{b}\in\mathbb{R}^{d_{\mathrm{out}}}$ (output bias).\medskip

Fixed parameter: \(P\in\mathbb{R}^{d_{\mathrm{in}}\times d_{\mathrm{out}}}\) is a non-trainable random projection matrix, initialized using one of the following schemes:
\begin{itemize}[itemsep=0.3ex, leftmargin=*]
    \item Gaussian: \(P_{ij}\sim\mathcal{N}\!\big(0,\tfrac{1}{d_{\mathrm{in}}}\big)\).
    \item Sparse ternary: \(P_{ij}\in\{-\alpha,0,+\alpha\}\) with equal probability, where \(\alpha=\sqrt{\tfrac{3}{d_{\mathrm{in}}}}\).
    \item Orthogonal: \(P\) has orthonormal columns (\(P^\top P = I\)). It is constructed by sampling a Gaussian matrix \(G \in \mathbb{R}^{d_{\mathrm{in}}\times d_{\mathrm{out}}}\) and extracting the orthonormal factor (e.g., via QR decomposition).
\end{itemize}

\subsubsection{Random Matrix Guarantees}

Matrix initialization is a crucial practical choice because the non-trainable projection matrix $P$ fundamentally dictates the subspace where adaptation occurs. The effectiveness of this projection is fundamentally rooted in the Johnson–Lindenstrauss (J-L) Lemma \citep{Johnson1984ExtensionsOL}. This lemma guarantees that a fixed, high-dimensional random projection matrix $P$ can approximately preserve the geometric distances between data points when they are mapped to a lower-dimensional subspace.\medskip

Initialization schemes like Gaussian or Sparse Ternary \citep{Li2006VerySR} leverage the theoretical guarantees of the J-L lemma, ensuring the random projection preserves critical geometric distances in the input data. The use of Orthogonal initialization goes a step further, guaranteeing that the projection operation $\mathbf{x} \to P^\top \mathbf{x}$ is an isometry that further prevents the distortion or collapse of information norms during the initial mapping.

\subsubsection{Expressivity}
Although the random projection $P$ is fixed, the PRP layer remains expressive due to the learnable modulation parameters. Formally, the effective linear transformation is:
\begin{equation}
W_{\text{PRP}} = \operatorname{diag}(\mathbf{w})\, P^\top\, \operatorname{diag}(\boldsymbol{\alpha})
\end{equation}

This structured form corresponds to a rank-constrained operator whose effective rank is bounded by $\min(d_{\text{in}}, d_{\text{out}})$. Since high-dimensional random projections are typically full-rank, $W_{\text{PRP}}$ spans a large, dense subspace of possible linear mappings.
\subsubsection{Parameter and Compute Complexity}
\label{sec:compute}
The number of trainable parameters is reduced to $d_{\text{in}} + 2d_{\text{out}}$, 
in contrast to the standard FC layer, which requires $d_{\text{in}} \cdot d_{\text{out}} + d_{\text{out}}$ parameters. In memory-constrained settings or for projections across large dimensions, this is a major advantage.\medskip

The naive dense matrix multiplication with the fixed projection $P$ still costs 
$\mathcal{O}(d_{\text{in}} \cdot d_{\text{out}})$ operations for both the forward and backward pass. Hence compute savings are realized primarily in the number of learned parameters and their gradients (training memory and optimizer state), not necessarily in raw multiply-accumulate FLOPs unless $P$ is chosen to be structured or sparse.\medskip

Because $P$ is non-trainable, it can be stored in low precision, generated on the fly from a PRNG seed or held as an implicit structured operator, further reducing memory or computation footprint.

\section{Experiments}
\label{sec:experiments}
We evaluate the proposed Parametrized Random Projection (PRP) layer across a spectrum of task complexities, progressing from controlled synthetic environments to standard real-world benchmarks. The goal of this evaluation is to isolate the impact of PRP on model efficiency, training stability, and predictive performance. \medskip

For each experimental setting, we compare the proposed PRP architecture against a standard fully connected baseline, and for medium-scale tasks we also include a Low-Rank FC baseline. For large-scale tasks, however, the high input dimensionality makes it impossible to construct a Low-Rank FC model with a parameter count even remotely comparable to the PRP layer; its required rank would have to be reduced to impractically small values, causing model collapse and representational failure. In these scenarios, we focus exclusively on PRP and FC comparisons, which already highlights the structural efficiency and expressivity advantages of the proposed layer.
\subsection{Benchmarks and Datasets}
We use several standard benchmarks for our experiments: MNIST \citep{LeCun2005TheMD}, 
Fashion-MNIST \citep{xiao2017fashionmnistnovelimagedataset}, 
CIFAR-10 \citep{Krizhevsky2009LearningML}, and TinyImageNet \citep{Le2015TinyIV}. 
\begin{center}
\captionof{table}{Summary of Datasets Used for Benchmarking Experiments}
\vspace{0.3cm}
\label{tab:datasets}
\footnotesize
\setlength{\tabcolsep}{4pt}
\renewcommand{\arraystretch}{1.2}
\begin{tabular}{l c c c c}
\midrule
\textbf{Dataset} & \textbf{Train/Test} & \textbf{In Dim.} & \textbf{Out Dim.} & \textbf{Task} \\
\midrule
\multicolumn{5}{l}{\textbf{Synthetic Tasks}}\\
Linear & 200 / -- & 2 & 1 & Bin. Cls. \\
XOR & 400 / -- & 2 & 1 & Bin. Cls. \\
Checkerboard & 800 / -- & 2 & 1 & Bin. Cls. \\
Polynomial & 400 / -- & 1 & 1 & Regression \\
\midrule
\multicolumn{5}{l}{\textbf{Medium-scale Tasks}}\\
MNIST & 60K / 10K & 784 & 10 & Cls. \\
Fashion-MNIST & 60K / 10K & 784 & 10 & Cls. \\
MNIST (AE) & 60K / 10K & 784 & 784 & Recon. \\
\midrule
\multicolumn{5}{l}{\textbf{Large-scale Tasks}}\\
CIFAR-10 & 50K / 10K & $32\times32\times3$ & 10 & Cls. \\
TinyImageNet & 100K / 10K & $64\times64\times3$ & 200 & Cls. \\
\bottomrule
\end{tabular}
\end{center}

\subsection{Model Architectures}
For all experiments the PRP variant mirrors the baseline architecture exactly (in terms of layer depths and widths), but replaces every trainable linear layer with a Parametrized Random Projection layer. In these layers, the projection matrix $P$ is fixed (non-trainable) and initialized using a standard Gaussian distribution, as it was experimentally determined to be the most stable one, with the exception of the Autoencoder, which explicitly utilizes orthogonal initialization. As for the activation function, ReLU \citep{agarap2019deeplearningusingrectified} is used for all hidden layers across all models unless specified otherwise.
\subsubsection{Synthetic Data Benchmarks}
For the controlled experiments on synthetic data, we utilized simple feed-forward Multilayer Perceptrons (MLPs):
\begin{itemize}[itemsep=0.5ex, leftmargin=0pt]
    \item[] Linear Classification: A single-layer model with no hidden units ($2 \to 1$) and an identity function output.
    \item[] XOR Classification: A two-layer network with one hidden layer of width 16 ($2 \to 16 \to 1$).
    \item[] Concentric Circles Classification: Again, a two-layer network with one hidden layer of width 16 ($2 \to 16 \to 1$).
    \item[] Polynomial Regression: A regression network with two hidden layers of width 64 ($1 \to 64 \to 64 \to 1$) and a linear output.
\end{itemize}

\subsubsection{Image Classification MLPs (MNIST \& FashionMNIST)}
We employed a three-layer MLP accepting flattened $28 \times 28$ inputs for standard classification benchmarks:
\begin{itemize}[itemsep=0.5ex, leftmargin=0pt]
    \item[] Standard \& PRP Variants: These models use layer dimensions of $784 \to 512 \to 256 \to 10$.
    \item[] Low Rank (FC): To control for total parameter count, we compared against a bottlenecked FC baseline. This model restricts the first layer's rank significantly, using dimensions of $784 \to 4 \to 256 \to 10$. 
\end{itemize}

\subsubsection{Autoencoders}
We evaluated reconstruction on MNIST using a symmetrical autoencoder with the following structure:
\begin{itemize}[itemsep=0.5ex, leftmargin=0pt]
    \item[] Architecture: The encoder maps the input to a latent space via $784 \to 512 \to 512$. The decoder mirrors this with $512 \to 512 \to 784$. The output layer uses a Sigmoid activation.
    
    \item[] Low-Rank Baseline: To lower the parameter budget this baseline utilizes a bottlenecked structure, mapping $784 \to 10 \to 256$ in the encoder and $256 \to 10 \to 784$ in the decoder.
    
    \item[] PRP Specifics: The encoder utilizes orthogonal random projections. The decoder utilizes the transpose of the encoder's fixed matrices for its projections. 
    The decision to use the transposed fixed projection matrix ($P_{\text{dec}} = P_{\text{enc}}^T$) is a practical requirement for the stability of the parameter-efficient PRP Autoencoder. With only minimal trainable parameters (scaling vectors and bias), using an independent random decoder basis ($P_{\text{dec}} \neq P_{\text{enc}}^T$) led to optimization collapse due to geometric misalignment. Enforcing $P_{\text{dec}} = P_{\text{enc}}^T$ creates that necessary geometric alignment, stabilizing training and allowing the few parameters to effectively calibrate feature magnitudes for reconstruction.

\end{itemize}

\subsubsection{Convolutional Neural Networks (CIFAR-10 \& TinyImageNet)}
We adopted custom VGG-style \citep{simonyan2015deepconvolutionalnetworkslargescale}  backbones with Batch Normalization \citep{ioffe2015batchnormalizationacceleratingdeep}. The convolutional feature extractors are standard and fully trainable in all experiments; modifications were applied strictly to the classifier heads.
\begin{itemize}[itemsep=0.5ex, leftmargin=0pt]
    \item[] VGG Backbones: We utilized custom VGG architectures adapted for each dataset. For CIFAR-10, the backbone consists of three blocks (filter widths 32, 64, 128) producing 2,048 flattened features. For TinyImageNet, it consists of four blocks (filter widths 64, 128, 256, 512) producing 8,192 flattened features.
    \item[] Classifier Heads: The flattened features are processed by 3-layer MLPs with Dropout. The CIFAR-10 head uses dimensions $2048 \to 512 \to 256 \to 10$ with $p=0.5$ dropout. The TinyImageNet head uses wider dimensions $8192 \to 1024 \to 1024 \to 200$ with $p=0.2$ dropout.
\end{itemize}

\subsection{Training and Optimization}
All models were trained using the Adam optimizer \citep{kingma2017adammethodstochasticoptimization}. We applied an ExponentialLR decay schedule to stabilize convergence, with the decay rate $\gamma$ set according to the domain. For synthetic tasks, we used a slow decay with $\gamma = 0.9999$ to support prolonged training. For all other regimes we used a standardized decay rate of $\gamma = 0.95$.
\subsection{Batch Sizes and Training Duration}
Batch sizes were also standardized across experiments. For synthetic benchmarks, we utilized full-batch training (using the entire dataset per update step). For all other models, the batch size was fixed at 256. Training epochs were set to ensure convergence and no early-stopping was used:
\begin{itemize}[itemsep=0.5ex, leftmargin=0pt]
    \item[] Synthetic Benchmarks: Training duration varied by task complexity: 100 epochs (Linear), 3,000 epochs (XOR, Concentric Circles), and 4,000 epochs (Polynomial Regression).
    \item[] Standard MLP \& Autoencoder: Trained for 10 epochs (MNIST) or 20 epochs (FashionMNIST, Autoencoder).
    \item[] CNN Benchmarks: Trained for 30 epochs (CIFAR-10, TinyImageNet).
\end{itemize}
\subsection{Learning Rate Selection}
To ensure a rigorous comparison, we implemented a two-stage training protocol. In the first stage, we conducted a Learning Rate Range Test \citep{smith2017cyclicallearningratestraining} sweeping over a wide magnitude (typically $10^{-4}$ to $10$) to identify the optimal initial learning rate for each specific model architecture (Standard, PRP, and Low-Rank). We observed that PRP and Low-Rank architectures often require different learning rates compared to standard linear layers to achieve convergence. In the second stage, these optimal rates were fixed and used for the final multi-seed evaluation runs, ensuring that every model variant was trained in its optimal regime.
\subsection{Preprocessing}
For MNIST/FashionMNIST, inputs were flattened and normalized to $\mu = 0.5, \sigma = 0.5$. For CIFAR-10, standard augmentations were applied, including random cropping ($32\times32$ with padding 4) and horizontal flipping, followed by normalization. For TinyImageNet, heavy augmentation was used, including random resized cropping ($64\times64$), horizontal flipping, and color jittering ($0.4$), followed by normalization.
\subsection{Objectives and Evaluation.}
Loss functions were selected based on the task: Binary Cross-Entropy with Logits for binary synthetic classification, Cross-Entropy Loss for multi-class image classification, and Mean Squared Error (MSE) for regression and reconstruction. For performance evaluation, we report classification metrics (Top-1 Accuracy and Macro-F1 score) and regression/reconstruction metrics (MSE, Mean Absolute Error (MAE), and Coefficient of Determination $R^2$).
\subsection{Reporting Protocol}
To account for initialization stochasticity, all reported metrics are the mean and standard deviation calculated over three independent random seeds. Training and test loss curves for a representative run of each task are visualized to compare convergence dynamics. 

\section{Results}
\label{sec:results}
\subsubsection*{Synthetic benchmarks}
\begin{center}
    \begin{minipage}{\linewidth}
        \centering
        \captionof{table}{Performance and Parameter Count on Synthetic Classification Benchmarks.}
        \vspace{0.3cm}
        \label{tab:synthetic_cls}
        \footnotesize
        \setlength{\tabcolsep}{4pt}
        \renewcommand{\arraystretch}{1.2}
        \begin{tabular}{l c c c c}
            \midrule
            \textbf{Task} & \multicolumn{2}{c}{\textbf{Parameters}} & \multicolumn{2}{c}{\textbf{Accuracy}} \\
            \cmidrule(lr){2-3} \cmidrule(lr){4-5}
            & \textbf{Standard} & \textbf{PRP} & \textbf{Standard} & \textbf{PRP} \\
            \midrule
            Linear & 3 & 4 & $1.000$ & $1.000$ \\
            XOR & 65 & 52 & $1.000$ & $1.000$ \\
            Concentric Circles & 65 & 52 & $1.000$ & $1.000$ \\
            \bottomrule
        \end{tabular}
        \par
        \vspace{2ex} 
        \raggedright
        \textit{Note: Losses and standard deviations were negligible and are therefore excluded.}
        
    \end{minipage} 
\end{center}

\subsubsection*{Polynomial Regression}

\begin{center}
\captionof{table}{Results for the Polynomial Regression Task.}
\vspace{0.3cm}
\label{tab:synthetic_regression}
\footnotesize
\setlength{\tabcolsep}{4pt}
\renewcommand{\arraystretch}{1.2}
\begin{tabular}{l c c}
    \toprule
    \textbf{Metric} & \textbf{Standard} & \textbf{PRP} \\
    \midrule
    Parameters & 4{,}353 & 387 \\
    MSE (Loss) & $0.080 \pm 0.001$ & $0.083 \pm 0.003$ \\
    MAE & $0.222 \pm 0.001$ & $0.228 \pm 0.004$ \\
    R$^{2}$ & $0.99887 \pm 0.00001$ & $0.99883 \pm 0.00004$ \\
    \bottomrule
\end{tabular}
\end{center}

\subsubsection*{MNIST}
\begin{center}
\captionof{table}{MNIST Classification Results.}
\vspace{0.3cm}
\label{tab:mnist_mlp}
\scriptsize
\setlength{\tabcolsep}{3pt}
\renewcommand{\arraystretch}{1.2}
\begin{tabular}{l c c c}
\toprule
\textbf{Metric} &
\textbf{PRP} &
\textbf{Standard (FC)} &
\textbf{Low-Rank (FC)} \\
\midrule
Parameters &
3{,}108 &
535{,}818 &
6{,}990 \\
Accuracy (\%) &
$91.66 \pm 0.60$ &
$97.79 \pm 0.21$ &
$55.78 \pm 18.02$ \\
Macro-F1 &
$0.916 \pm 0.006$ &
$0.978 \pm 0.002$ &
$0.531 \pm 0.196$ \\
Train Loss &
$0.261 \pm 0.006$ &
$0.030 \pm 0.001$ &
$1.161 \pm 0.428$ \\
Test Loss (Final) &
$0.269 \pm 0.011$ &
$0.076 \pm 0.008$ &
$1.179 \pm 0.427$ \\
Best Test Loss &
$0.259 \pm 0.006$ &
$0.071 \pm 0.003$ &
$1.165 \pm 0.427$ \\
\bottomrule
\end{tabular}
\end{center}

\begin{center}
\includegraphics[width=\linewidth]{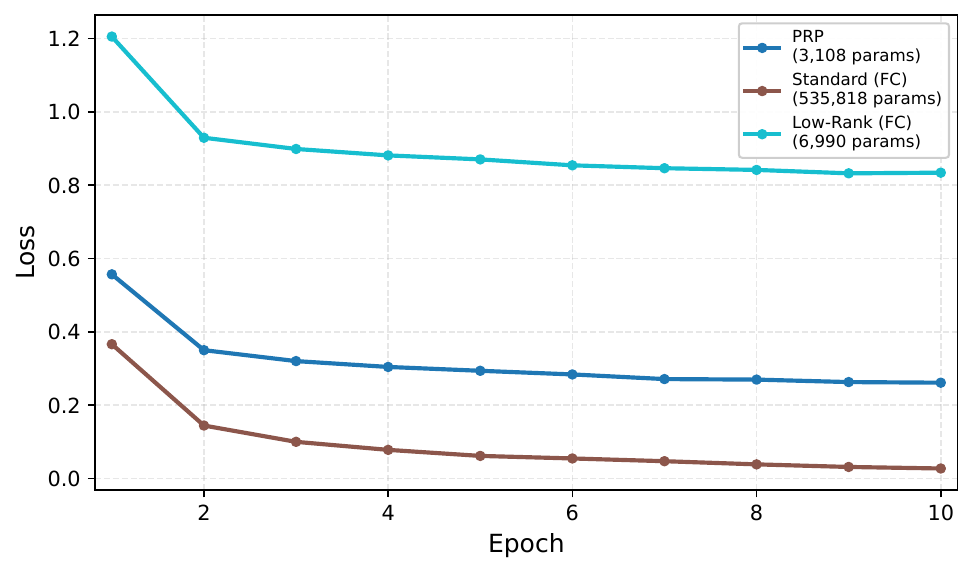}
\captionof{figure}{Training Loss on MNIST Classification.}
\label{fig:mnist_train}
\end{center}

\begin{center}
\includegraphics[width=\linewidth]{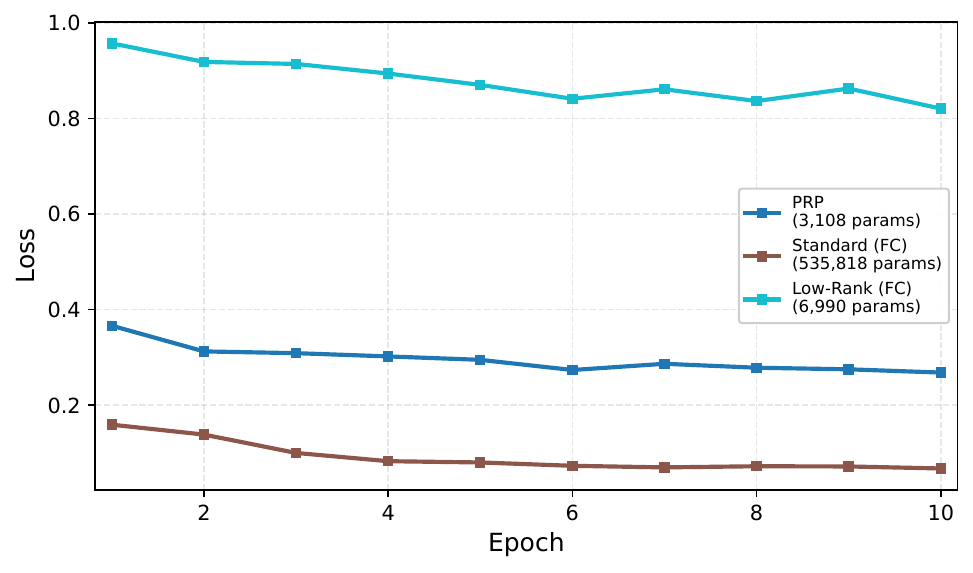}
\captionof{figure}{Test Loss on MNIST Classification.}
\label{fig:mnist_val}
\end{center}

\subsubsection*{Fashion-MNIST}
\begin{center}
\captionof{table}{Fashion-MNIST Classification Results.}
\vspace{0.3cm}
\label{tab:fmnist_mlp}
\scriptsize
\setlength{\tabcolsep}{4pt}
\renewcommand{\arraystretch}{1.2}
\begin{tabular}{l c c c}
\textbf{Metric} &
\textbf{PRP} &
\textbf{Standard (FC)} &
\textbf{Low-Rank (FC)} \\
\midrule
Parameters &
3{,}108 &
535{,}818 &
6{,}990 \\
Accuracy (\%) &
$83.78 \pm 0.52$ &
$89.33 \pm 0.33$ &
$82.59 \pm 0.13$ \\
Macro-F1 &
$0.838 \pm 0.003$ &
$0.893 \pm 0.003$ &
$0.825 \pm 0.003$ \\
Train Loss &
$0.386 \pm 0.002$ &
$0.108 \pm 0.002$ &
$0.431 \pm 0.004$ \\
Test Loss (Final) &
$0.449 \pm 0.009$ &
$0.368 \pm 0.004$ &
$0.493 \pm 0.004$ \\
Best Test Loss &
$0.440 \pm 0.001$ &
$0.327 \pm 0.002$ &
$0.493 \pm 0.004$ \\
\bottomrule
\end{tabular}
\end{center}

\begin{center}
\includegraphics[width=\linewidth]{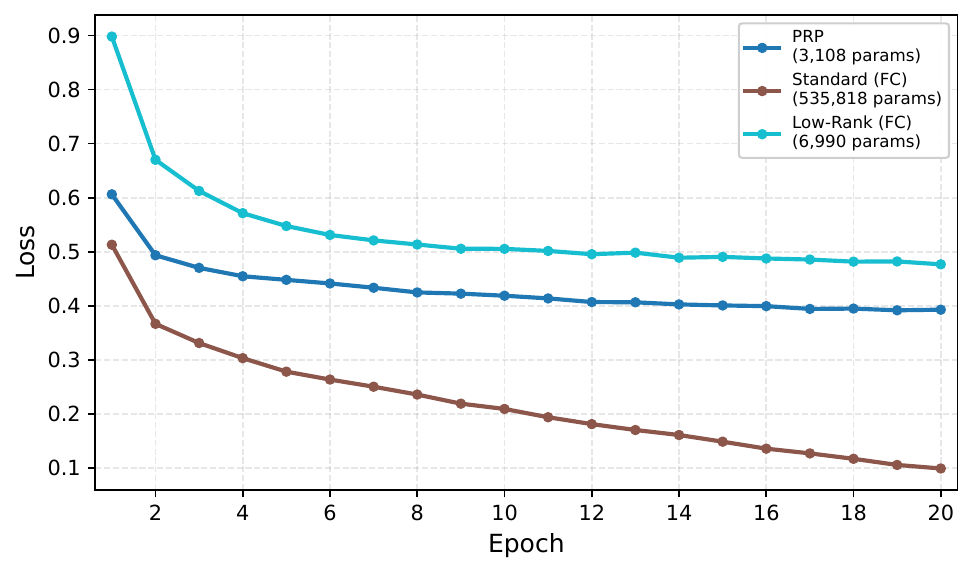}
\captionof{figure}{Training Loss on FMNIST Classification.}
\label{fig:fmnist_train}
\end{center}

\begin{center}
\includegraphics[width=\linewidth]{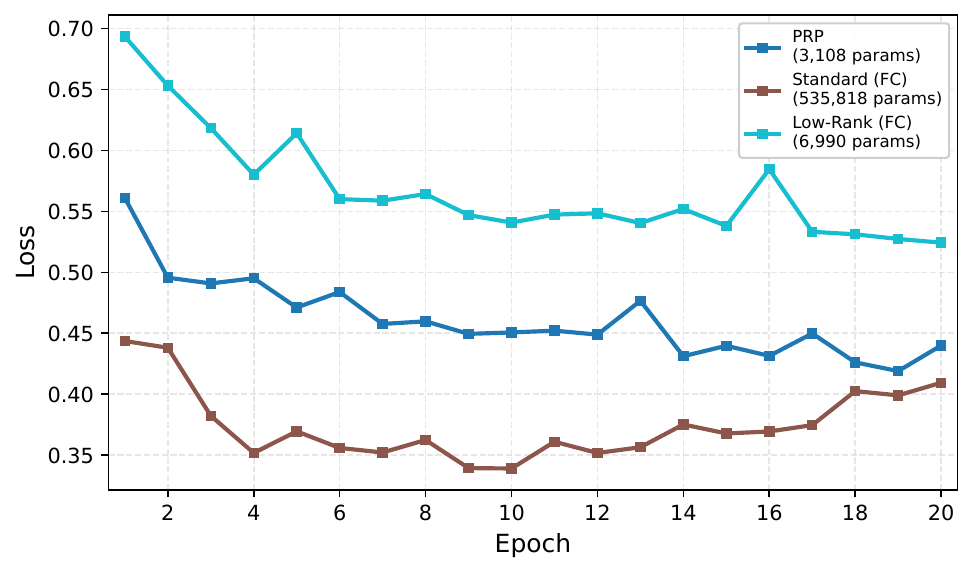}
\captionof{figure}{Test Loss on FMNIST Classification.}
\label{fig:fmnist_val}
\end{center}

\subsubsection*{MNIST Autoencoder}
\noindent\begin{minipage}{\linewidth}
    \centering
    \captionof{table}{MNIST Autoencoder Reconstruction Performance.}
    \vspace{0.3cm}
    \label{tab:mnist_ae}
    \scriptsize
    \setlength{\tabcolsep}{4pt}
    \renewcommand{\arraystretch}{1.2}
    \begin{tabular}{l c c c}
        \textbf{Metric} &
        \textbf{PRP} &
        \textbf{Standard (FC)} &
        \textbf{Low Rank (FC)} \\
        \midrule
        Parameters &
        6{,}960 &
        1{,}329{,}424 &
        21{,}860 \\
        MSE &
        $0.0165 \pm 0.0003$ &
        $0.0034 \pm 0.0001$ &
        $0.0675 \pm 0.0001$ \\
        MAE &
        $0.0497 \pm 0.0006$ &
        $0.0170 \pm 0.0005$ &
        $0.1522 \pm 0.0006$ \\
        Test Loss &
        $0.0165 \pm 0.0003$ &
        $0.0034 \pm 0.0001$ &
        $0.0675 \pm 0.0001$ \\
        Best Test Loss &
        $0.0165 \pm 0.0003$ &
        $0.0034 \pm 0.0001$ &
        $0.0675 \pm 0.0001$ \\
        \bottomrule
    \end{tabular}
\end{minipage}

\begin{center}
    \includegraphics[width=0.9\linewidth]{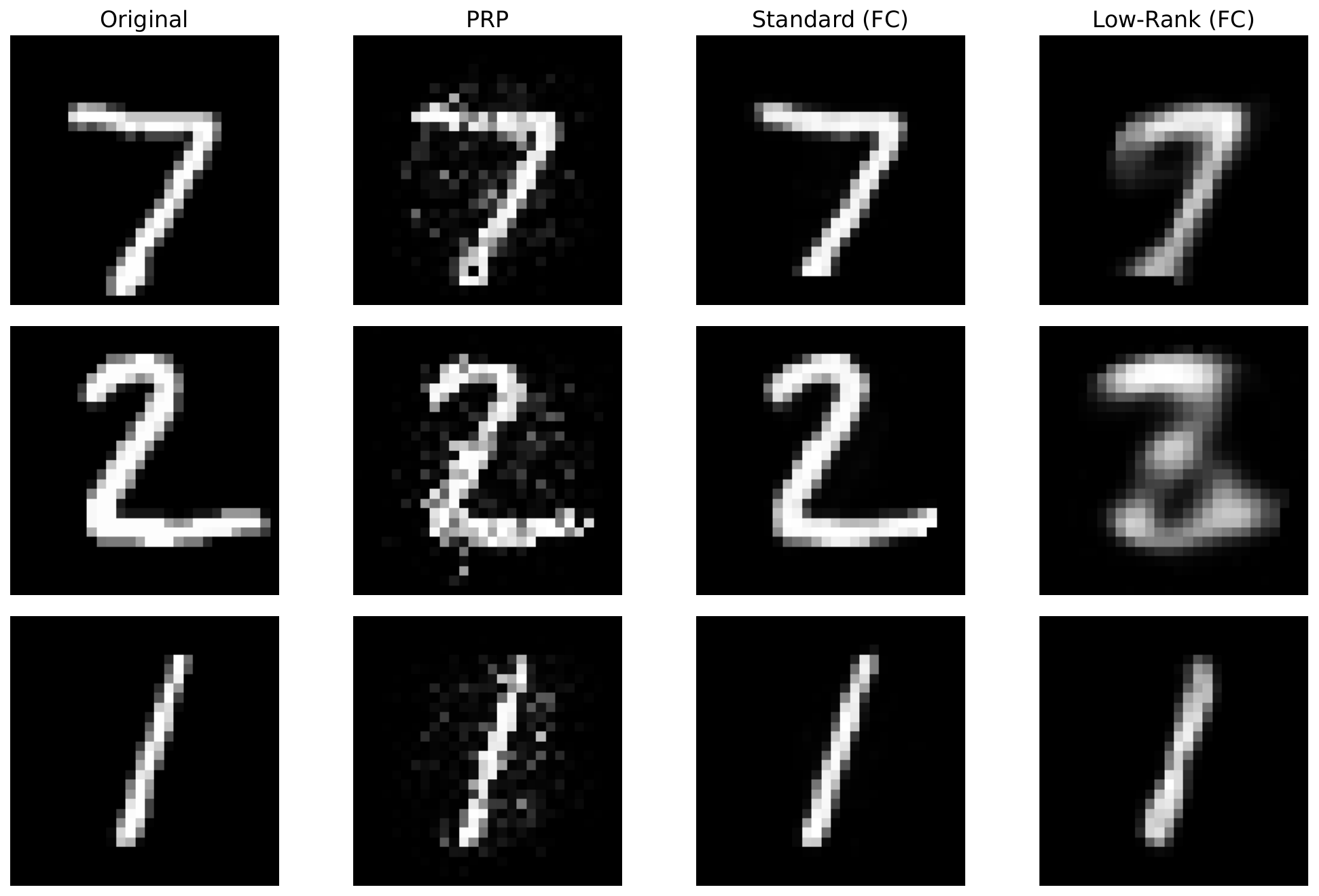}
    \captionof{figure}{Reconstruction Samples from the MNIST Autoencoder. First column shows original images.}
    \label{fig:mnist_ae_reconstruction}
\end{center}
\begin{center}
    \includegraphics[width=\linewidth]{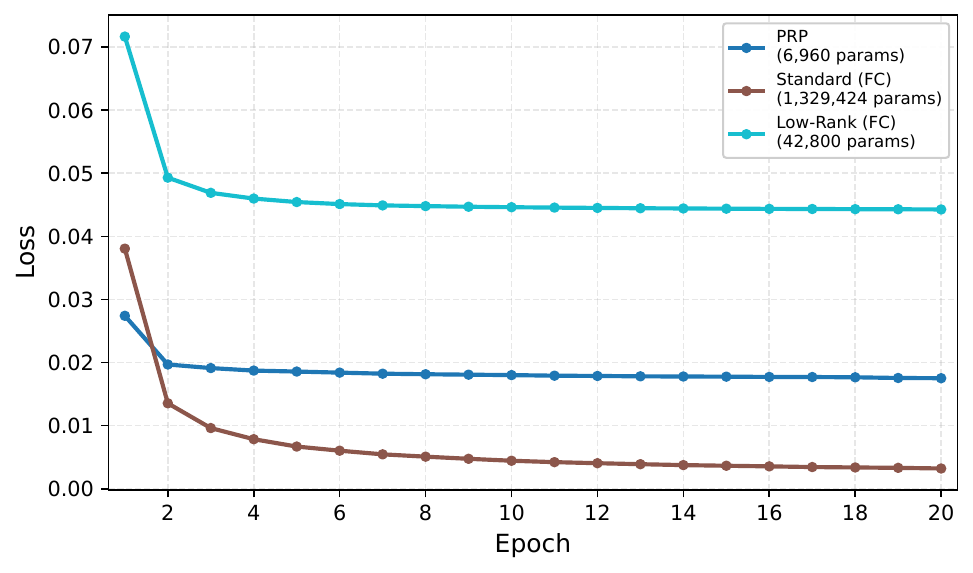}
    \captionof{figure}{Training Loss for the MNIST Autoencoder.}
    \label{fig:mnist_ae_train_loss}
\end{center}
\begin{center}
    \includegraphics[width=\linewidth]{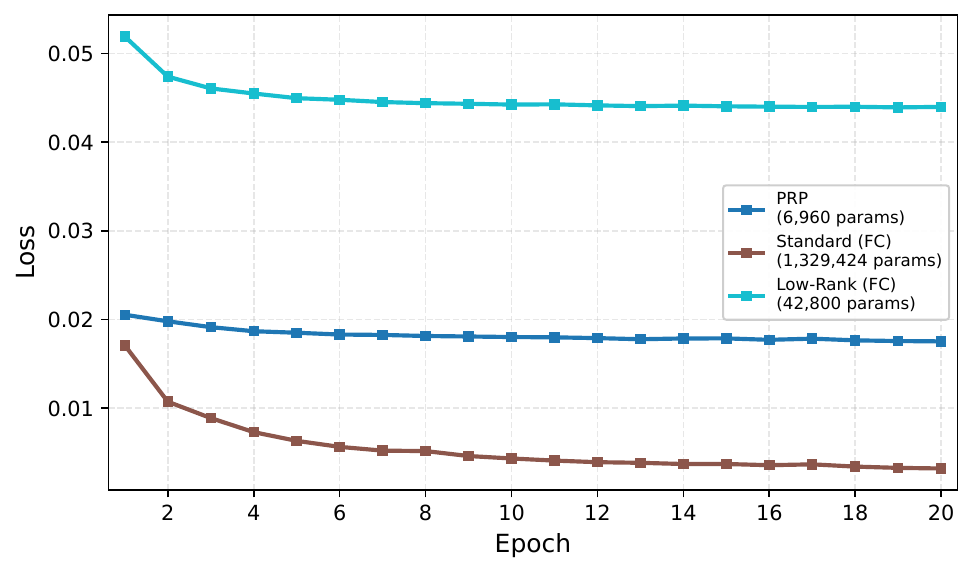}
    \captionof{figure}{Test Loss for the MNIST Autoencoder. }
    \label{fig:mnist_ae_val_loss}
\end{center}

\subsubsection*{CIFAR-10}
\begin{center}
\captionof{table}{CIFAR-10 Classification Results (VGG).}
\vspace{0.3cm}
\label{tab:cifar10_vgg}
\footnotesize
\setlength{\tabcolsep}{4pt}
\renewcommand{\arraystretch}{1.2}
\begin{tabular}{l c c}
\toprule
\textbf{Metric} &
\textbf{PRP VGG} &
\textbf{Standard VGG} \\
\midrule
Classifier Params &
4{,}372 &
1{,}182{,}986 \\
Total Params &
292{,}276 &
1{,}470{,}890 \\
Accuracy (\%) &
$87.40 \pm 0.26$ &
$86.21 \pm 0.44$ \\
Macro-F1 &
$0.874 \pm 0.003$ &
$0.861 \pm 0.005$ \\
Train Loss &
$0.293 \pm 0.002$ &
$0.308 \pm 0.001$ \\
Test Loss (Final) &
$0.394 \pm 0.004$ &
$0.413 \pm 0.004$ \\
Best Test Loss &
$0.389 \pm 0.003$ &
$0.411 \pm 0.003$ \\
\bottomrule
\end{tabular}
\end{center}

\vspace{0.7cm}
\begin{center}
    \includegraphics[width=\linewidth]{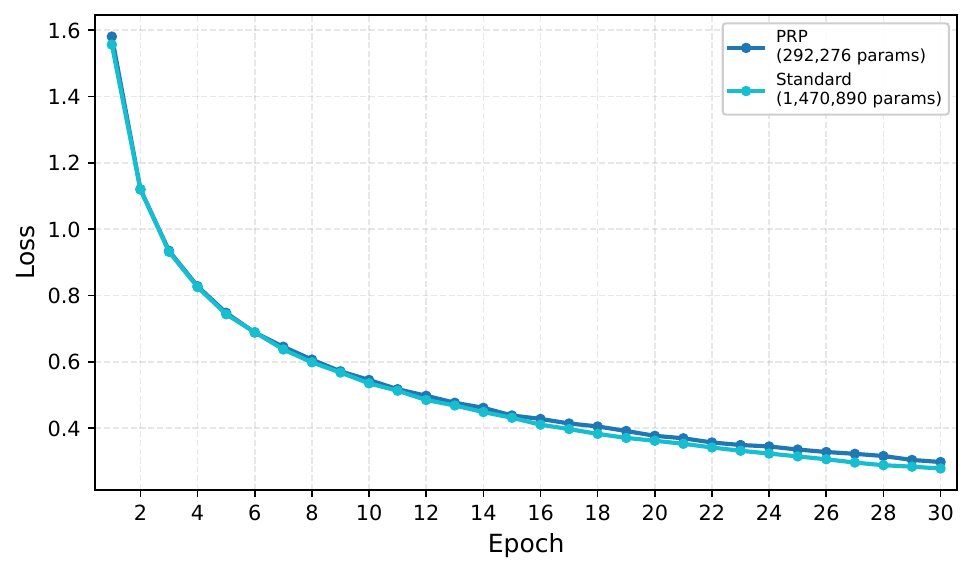}
    \captionof{figure}{Training Loss on CIFAR-10 VGG Backbone.}
    \label{fig:cifar10_train_loss}
\end{center}

\begin{center}
    \includegraphics[width=\linewidth]{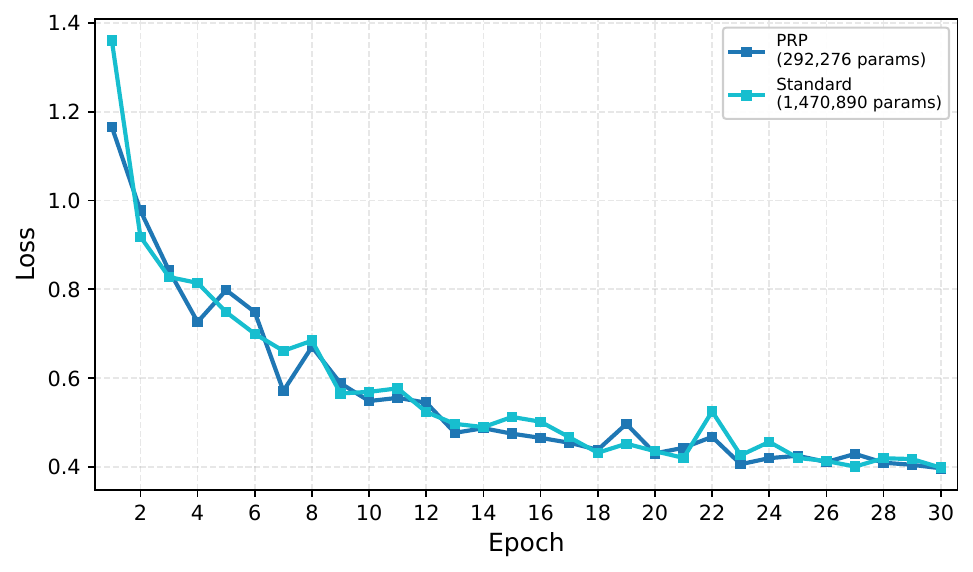}
    \captionof{figure}{Test Loss on CIFAR-10 VGG Backbone.}
    \label{fig:cifar10_val_loss}
\end{center}

\subsubsection*{TinyImageNet}
\noindent\begin{minipage}{\linewidth}
    \centering
    \captionof{table}{TinyImageNet Classification Results.}
    \vspace{0.3cm}
    \label{tab:tinyimagenet}
    \footnotesize
    \setlength{\tabcolsep}{4pt}
    \renewcommand{\arraystretch}{1.2}
    \begin{tabular}{l c c}
        \toprule
        \textbf{Metric} &
        \textbf{PRP VGG} &
        \textbf{Standard VGG} \\
        \midrule
        Classifier Params &
        14{,}736 &
        9{,}644{,}232 \\
        Total Params &
        5{,}590{,}224 &
        15{,}219{,}720 \\
        Accuracy (\%) &
        47.44 &
        45.84 \\
        Macro-F1 &
        0.467 &
        0.453 \\
        Train Loss &
        1.746 &
        1.512 \\
        Test Loss (Final) &
        2.309 &
        2.342 \\
        Best Test Loss &
        2.272 &
        2.262 \\
        \bottomrule
    \end{tabular}
    \par
    \vspace{2ex} 
    \raggedright
    \textit{Note: Due to computational constraints, TinyImageNet benchmarks were evaluated in a single run; therefore, mean and standard deviation are not reported.}
\end{minipage}
\vspace{0.5cm}

\begin{center}
    \includegraphics[width=\linewidth]{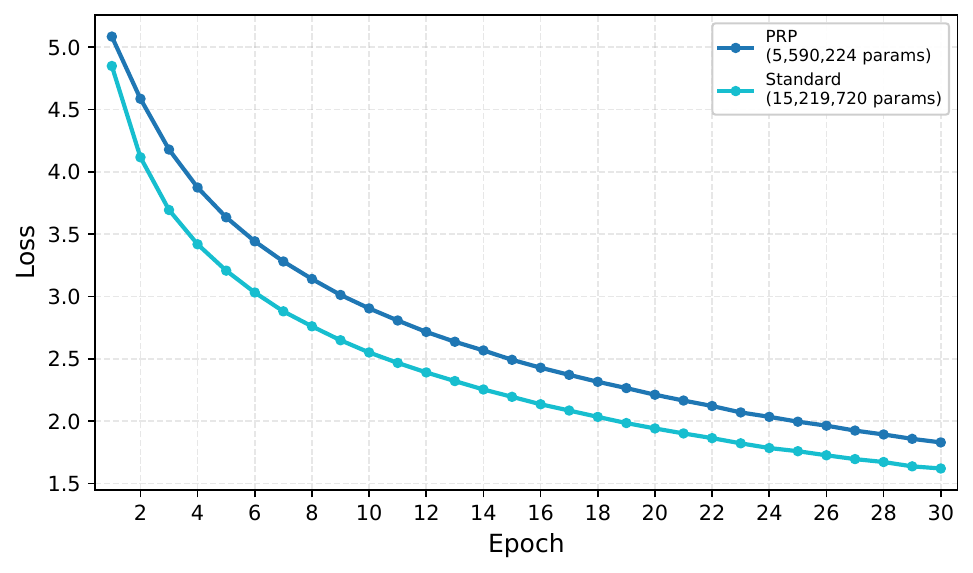}
    \captionof{figure}{Training Loss on TinyImageNet.}
    \label{fig:timagenet_train_loss}
\end{center}
\begin{center}
    \includegraphics[width=\linewidth]{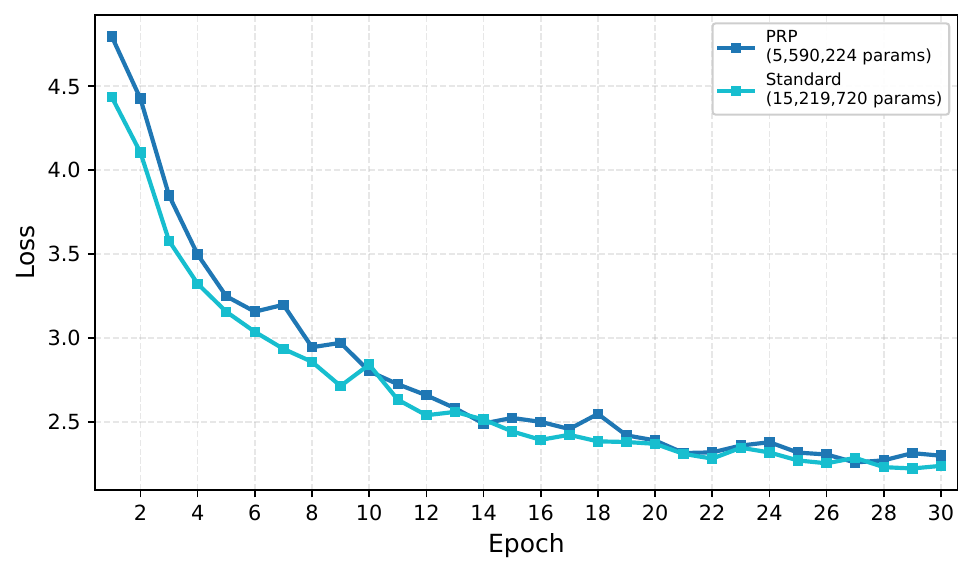}
    \captionof{figure}{Test Loss on TinyImageNet.}
    \label{fig:timagenet_val_loss}
\end{center}

\section{Discussion}
\subsection{Analysis of Parameter Efficiency and Generalization}
Using synthetic tasks, where the PRP and Standard models went head-to-head in performance (\cref{tab:synthetic_cls})(\cref{tab:synthetic_regression}) allowed us to  demonstrate its equal capability under regimes where simple memorization and feature mapping suffice, thus proving its basic representational capabilities in low-complexity settings. Moving to more realistic domains, the PRP layer redefines the efficiency trade-off: extreme parameter savings (e.g., $172\times$ fewer parameters in \cref{tab:mnist_mlp}) result in minimal performance reduction, retaining more than 93\% of the baseline accuracy. This efficiency is most pronounced in the autoencoder task, where the PRP model achieved a $191\times$ parameter reduction (\cref{tab:mnist_ae}) relative to the Standard model. Moreover, the proposed model outperformed the Low-Rank reference architecture in reconstruction quality while using only one-third of its parameters. Notably, on the large-scale image recognition tasks, CIFAR-10 (\cref{tab:cifar10_vgg}) and TinyImageNet (\cref{tab:tinyimagenet}), the PRP VGG architecture matched or exceeded the performance of the baseline models in terms of classification accuracy, despite a total parameter reduction (5$\times$ and 2.72$\times$ respectively) and a significant decrease in the classifier head's parametrization.
\subsubsection*{Representational Density Analysis}

To formally quantify the efficiency of the PRP layer, we introduce the Bit Efficiency Score ($\mathcal{E}_{\text{BES}}$). Unlike simple parameter ratios, which unfairly penalize necessary model scaling and fail to account for increasing task complexity and model expansion, this metric serves as a direct measure of capacity utilization. It weighs the performance gains of the model against the logarithmic scale of its parameters, highlighting how much predictive value each order of magnitude in model capacity contributes relative to the inherent complexity of the task.

We calculate the Bit Efficiency Score ($\mathcal{E}_{\text{BES}}$) as:
\begin{equation}
\mathcal{E}_{\text{BES}} = \mathbf{\mathcal{P}} \times \frac{\log_2(\mathbf{N} \times \mathbf{D}_{\text{in}})}{\log_2(\mathbf{\theta})}
\end{equation}
Where:\\
$\mathbf{\mathcal{P}}$ is the Performance Gain, $\mathbf{N} \times \mathbf{D}_{\text{in}}$ represents the total information content of the training data (samples $\times$ input dimension), and $\mathbf{\theta}$ is the parameter count which represents the model's storage capacity.
\newline

The logarithmic ratio accounts for the exponential nature of information scaling. This ensures the metric remains stable and fair across tasks of varying dimensional magnitudes.

In the specific context of classification accuracy, where the baseline is a random guess:
$$
\mathbf{\mathcal{P}} = \text{Validation Accuracy} - \text{Chance Baseline}
$$

\noindent\begin{minipage}{\linewidth}
\centering
\captionof{table}{Bit Efficiency Score ($\mathcal{E}_{\text{BES}}$) results for chosen tasks.}
\vspace{0.3cm}
\label{tab:unified_data_model_efficiency}
\footnotesize
\setlength{\tabcolsep}{4pt}
\renewcommand{\arraystretch}{1.2}
\begin{tabular}{l l c c}
\toprule
\textbf{Task} & \textbf{Model} & \textbf{Parameters ($\mathbf{\theta}$)} & \textbf{$\mathbf{\mathcal{E}_{\text{BES}}}$ Score} \\
\midrule
MNIST & Standard (FC) & 535{,}818 & 1.18 \\
 & Low-Rank (FC) & 6{,}990 & 0.91 \\
 & PRP & 3{,}108 & \textbf{1.79} \\
\midrule
Fashion-MNIST & Standard (FC) & 535{,}818 & 1.06 \\
 & Low-Rank (FC) & 6{,}990 & 1.45 \\
 & PRP & 3{,}108 & \textbf{1.62} \\
\midrule
CIFAR-10 & Standard VGG & 1{,}470{,}890 & 1.01 \\
 & PRP VGG & 292{,}276 & \textbf{1.16} \\
\bottomrule
\end{tabular}
\end{minipage}

\subsection{Limitations and Practical Implications}
\subsubsection*{Computational Cost}
The PRP layer does not reduce raw computational throughput compared to a standard fully connected layer. As a result, no runtime gains are realized in terms of FLOPs. Efficiency improvements arise only in training memory, storage, and optimizer state reduction rather than in the speed of the matrix multiplications themselves. Meaningful FLOP savings would require $P$ to be structured or sparse.
\subsubsection*{Absolute Accuracy Ceiling} 
The PRP layer prioritizes efficiency and parameter reduction, and as such it is not competitive with dense layers when absolute accuracy is the primary objective. The non-trainable projection matrix restricts the transformation to a fixed random subspace, limiting the layer’s effective rank. Although high-dimensional random projections benefit from robustness properties such as Johnson–Lindenstrauss, they cannot match the expressiveness of a fully optimized dense operator. This structural rigidity prevents the PRP layer from representing the full spectrum of target linear mappings, thereby imposing an upper-bound on potential accuracy. 
\subsubsection*{Subspace Restriction} 
Because $P$ is fixed, the layer operates strictly within its predefined random subspace. While empirical results show that PRP is resilient to many initialization choices, theoretical guarantees remain weaker compared to fully trainable dense transformations. This constraint reduces representational freedom and affects the layer’s ability to adapt its geometric structure to complex tasks.
\subsection{Future Directions}
Future research should prioritize the exploration of hybrid architectures that strategically interleave standard dense layers with PRP modules. To enhance expressivity without compromising efficiency, this approach can be augmented by learnable low-rank residual paths parallel to the fixed projections. Furthermore, extending PRP methodology to convolutional operations offers a pathway for structured, parameter-efficient feature extraction in computer vision domains. From a computational standpoint, leveraging structured sparsity or Subsampled Randomized Hadamard Transforms could significantly reduce the FLOPS required for such layers. Finally, applying PRP to Transformer architectures: specifically within the large dense projection matrices $W_Q$, $W_K$, and $W_V$—presents a high-impact opportunity for scaling efficient attention mechanisms.

\section{Related work}
Research on random projections and randomized neural networks provides the closest precedents for our approach. Fixed random projection layers are commonly used as a preprocessing stage in deep models, often with Gaussian or Achlioptas-distributed matrices, to reduce input dimensionality and improve computational efficiency \citep{wójcik2018randomprojectiondeepneural}. Random projections have also been used to enhance expressive power and accelerate training; for instance, \citep{cai2019enhancedexpressivepowerfast} apply a fixed random matrix prior to a trainable fully connected layer to simplify subsequent learning. Similarly, Extreme Learning Machines (ELMs) \citep{Huang2006ExtremeLM} fix hidden layer weights entirely while training only the output layer, achieving fast training and strong empirical performance. Beyond randomization, alternative research directions improve efficiency by restructuring the weight matrix itself. Low-rank factorizations approximate weight matrices in dense (and convolutional) layers as the product of two smaller matrices, reducing parameter count and computational cost \citep{denton2014exploitinglinearstructureconvolutional}. However, this imposes an explicit rank constraint that can limit the model's expressive power. In contrast, PRP modulates a full-rank random projection, avoiding these constraints to outperform low-rank baselines. Finally, structured matrix approaches, such as those based on circulant or Toeplitz operators, exploit algebraic structure to allow for fast multiplication via algorithms like the FFT \citep{sindhwani2015structuredtransformssmallfootprintdeep, cheng2015explorationparameterredundancydeep}. While these methods reduce complexity to $\mathcal{O}(n \log n)$, they achieve savings through deterministic constraints rather than the statistical properties leveraged by our approach.

\section{Conclusion}
This work demonstrates that the dense parameterization of traditional projection layers is largely unnecessary for learning effective representations in high-dimensional spaces. By replacing the weight matrix with a fixed random subspace and isolating the adaptation process to simple scaling vectors, the PRP layer achieves parameter reductions of up to two orders of magnitude while preserving geometric properties and predictive performance. Our experiments confirm that this method serves as a robust replacement for standard projection layers, successfully retaining over 90\% of baseline accuracy even in extreme compression regimes. 

\newpage
\bibliographystyle{icml2025}
\bibliography{references}

\end{multicols}
\end{document}